\documentclass[11pt]{article}
\usepackage[a4paper,margin=2.2cm]{geometry}
\usepackage{graphicx}
\usepackage{booktabs}
\usepackage{array}
\usepackage{longtable}
\usepackage{amsmath,amssymb}
\usepackage{hyperref}
\usepackage{enumitem}
\usepackage{caption}
\usepackage{xcolor}
\usepackage{tikz}
\usetikzlibrary{arrows.meta,positioning,shapes.geometric}
\usepackage{tabularx}
\hypersetup{colorlinks=true,linkcolor=black,citecolor=black,urlcolor=blue}
\title{Fine-Tuning General-Purpose Large Language Models for Agricultural Applications:\\A Reproducible Framework and Evaluation Protocol Based on Qwen3-8B}
\author{Zhaoyang Li\textsuperscript{1,*} \quad Ruijie Zhang\textsuperscript{1} \quad Jiaqi Liu\textsuperscript{2} \quad Zhaoji Sun\textsuperscript{1}\\
\small \textsuperscript{1}Sanya University \\[-0.1em]
\small \textsuperscript{2}Hebei International Studies University \\[-0.1em]
\small \textsuperscript{*}Corresponding author: Zhaoyang Li, \href{mailto:beijizhi@outlook.com}{beijizhi@outlook.com}}
\date{}

\begin{document}
\maketitle

\begin{abstract}
General-purpose large language models (LLMs) have demonstrated strong abilities in open-domain question answering, information extraction, and text generation. Agricultural applications, however, are domain-specific, region-dependent, time-sensitive, and safety-critical. Without data governance, expert evaluation, and evidence constraints, an agricultural assistant may produce unreliable advice on crop diseases, pesticide use, fertilization, or policy interpretation. This paper does not report any model-performance claims that have not been produced by an actual training run and expert evaluation. Instead, we propose AgriTune-R, a reproducible and auditable framework for adapting general-purpose LLMs to agricultural tasks. The framework selects the publicly verifiable Qwen3-8B model as the recommended base model and integrates agricultural data governance, instruction construction, LoRA/QLoRA parameter-efficient fine-tuning, retrieval-augmented generation, expert evaluation, and safety control for high-risk questions. The contributions are: (1) a structured workflow for agricultural LLM adaptation; (2) an evaluation protocol for agricultural knowledge QA, pest and disease consultation, cultivation management, and policy explanation; (3) an expert-review rubric combining factuality, safety, evidence consistency, and uncertainty expression; and (4) a clear separation between protocol design and empirical conclusions, providing an executable baseline for future empirical studies.
\end{abstract}


\section{Introduction}
Agricultural decision-making relies on crop physiology, disease symptoms, soil conditions, weather, local policy, pesticide registration, and practical cultivation experience. Although general-purpose LLMs can process natural-language questions, their parametric knowledge is not guaranteed to contain current, local, or safety-compliant agricultural information. In farmer-facing consultation, a fluent but incorrect answer may lead to pesticide misuse, delayed disease control, economic loss, or food-safety risks. Consequently, agricultural LLM research should evaluate not only fluency but also data provenance, evidence chains, expert review, risk control, and deployment responsibility.

Recent studies suggest that agricultural foundation models and domain-specific LLM systems are promising, but they also face challenges related to data quality, benchmark design, hallucination, and deployment safety \cite{li2023smartagriculture,agrogpt2024,shizishan2024,agrigpt2025}. AgriBERT injects agricultural and food-related knowledge into language models for matching food descriptions and nutrition data \cite{agribert2022}. AgroGPT focuses on agricultural vision-language instruction construction and multimodal dialogue \cite{agrogpt2024}. ShizishanGPT integrates retrieval-augmented generation, knowledge graphs, and tool use for agricultural question answering \cite{shizishan2024}. AgriGPT further emphasizes data engines, domain-specific benchmarks, and multi-channel retrieval augmentation \cite{agrigpt2025}. These studies show that an agricultural assistant should be assessed not merely by whether it answers, but by whether its answer is grounded, safe, and reviewable.

This paper clarifies its research position. It is not a report claiming that a model has already been trained and has achieved numerical improvements. Instead, it is a methods and evaluation-protocol paper intended to provide a complete, executable, and auditable foundation for future empirical fine-tuning studies. To ensure model-source accuracy and reproducibility, we select Qwen3-8B as the recommended base model. The Qwen3 technical report and official repository list publicly available Qwen3 model scales including 0.6B, 1.7B, 4B, 8B, 14B, 32B, and MoE variants; therefore, the 8B-scale model is appropriate for a reproducible paper \cite{qwen3tech,qwen3github}.

The main contributions of this paper are as follows:
\begin{enumerate}[leftmargin=2em]
    \item We propose AgriTune-R, a framework that integrates agricultural data governance, instruction fine-tuning, retrieval-augmented generation, and safety evaluation.
    \item We provide a Qwen3-8B-based adaptation protocol while avoiding unverifiable model names and unexecuted performance numbers.
    \item We design an agricultural evaluation protocol covering knowledge QA, pest and disease consultation, cultivation management, policy explanation, and high-risk refusal.
    \item We present practical tables and execution rules for expert scoring, evidence consistency, risk control, and reproducibility logging.
\end{enumerate}

\section{Related Work}
\subsection{General LLMs and parameter-efficient adaptation}
General LLMs acquire language understanding, reasoning, and generation abilities through large-scale pretraining. Full fine-tuning can be expensive and data-demanding when such models are adapted to specialized domains. LoRA freezes pretrained weights and injects trainable low-rank matrices, reducing the number of trainable parameters \cite{lora2021}. QLoRA trains LoRA adapters through quantized base-model weights, further reducing memory requirements \cite{qlora2023}. These methods make agricultural adaptation feasible for universities, local extension services, and small research teams with limited computing resources.

\subsection{Retrieval-augmented generation}
Agricultural knowledge changes over time. Pesticide registration, local policy, pest and disease alerts, and weather-related recommendations may become outdated. Relying only on model parameters can therefore produce obsolete or unsupported answers. Retrieval-augmented generation (RAG) supplies external evidence to the generation model, making answers more traceable \cite{rag2020}. In agriculture, RAG is particularly useful for extension manuals, policy explanation, disease-control guidance, and safety-standard checking.

\subsection{Agricultural language and multimodal models}
Agricultural model research is moving from single-task models to domain foundation models and model ecosystems. AgriBERT addresses semantic matching in agricultural and food-related text \cite{agribert2022}. Reviews of smart-agriculture foundation models emphasize the need to connect language, vision, multimodal, and decision-making tasks while accounting for data quality and deployment risk \cite{li2023smartagriculture}. AgroGPT shows that expert-tuned agricultural vision-language data can reduce domain gaps \cite{agrogpt2024}. ShizishanGPT and AgriGPT highlight the importance of retrieval, knowledge graphs, tool use, and domain-specific evaluation in agricultural QA systems \cite{shizishan2024,agrigpt2025}.

\section{Model Selection and Truthfulness Principles}
Truthfulness is a prerequisite for the proposed study design. Any content that is not supported by an actual training run, inference experiment, or expert evaluation is not written as an empirical conclusion. Model selection follows three principles: public verifiability, community reproducibility, and feasible computing cost. Based on these principles, we recommend Qwen3-8B rather than an unverifiable model name.

\begin{table}[htbp]
\centering
\caption{Principles for selecting the base model}
\begin{tabularx}{\textwidth}{p{3.2cm}X}
\toprule
Item & Description \\
\midrule
Recommended base model & Qwen3-8B, a dense model scale listed in the public Qwen3 family and suitable for subsequent LoRA/QLoRA experiments. \\
Rationale & The 8B scale is more feasible for single-machine or small multi-GPU settings than larger models, and it is more reproducible than an unverifiable model name. \\
Not claimed & This paper does not claim that Qwen3-8B has already been fine-tuned for agriculture, nor does it report accuracy, F1, win rate, or any other number without real training and expert evaluation. \\
Executable next step & Researchers can execute the protocol in this paper and then add result tables and a model card after completing real experiments. \\
\bottomrule
\end{tabularx}
\end{table}

\section{The AgriTune-R Framework}
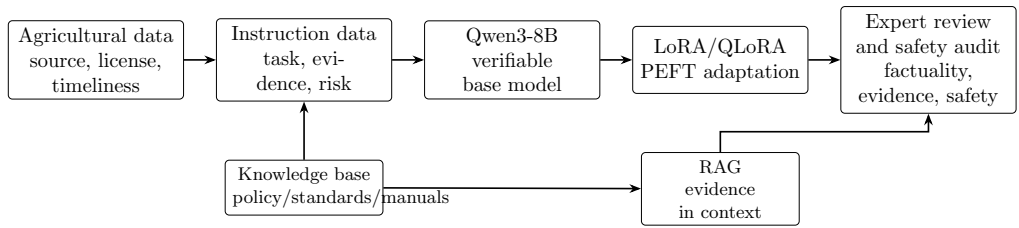
\begin{figure}[htbp]
\centering
\resizebox{0.98\textwidth}{!}{%
\begin{tikzpicture}[
    box/.style={draw, rounded corners=2pt, align=center, minimum height=1.15cm, text width=2.75cm, font=\small},
    smallbox/.style={draw, rounded corners=2pt, align=center, minimum height=0.9cm, text width=2.45cm, font=\footnotesize},
    arrow/.style={-{Stealth[length=2mm]}, thick}
]
\node[box] (data) {Agricultural data\\source, license, timeliness};
\node[box, right=0.55cm of data] (inst) {Instruction data\\task, evidence, risk};
\node[box, right=0.55cm of inst] (base) {Qwen3-8B\\verifiable base model};
\node[box, right=0.55cm of base] (peft) {LoRA/QLoRA\\PEFT adaptation};
\node[box, right=0.55cm of peft] (eval) {Expert review and safety audit\\factuality, evidence, safety};
\draw[arrow] (data) -- (inst);
\draw[arrow] (inst) -- (base);
\draw[arrow] (base) -- (peft);
\draw[arrow] (peft) -- (eval);
\node[smallbox, below=1.0cm of inst] (kb) {Knowledge base\\policy/standards/manuals};
\node[smallbox, below=1.0cm of peft] (rag) {RAG\\evidence in context};
\draw[arrow] (kb) -- (rag);
\draw[arrow] (kb.north) -- ++(0,0.35) -| (inst.south);
\draw[arrow] (rag.north) -- ++(0,0.35) -| (eval.south);
\node[align=center, font=\footnotesize, below=0.55cm of rag] {The figure is a workflow description only; no performance numbers are reported before real training and expert evaluation.};
\end{tikzpicture}%
}
\caption{Overview of AgriTune-R. The figure describes the method workflow and does not report experimental results.}
\label{fig:framework}
\end{figure}

AgriTune-R decomposes agricultural adaptation into auditable steps, addressing common problems such as unclear data provenance, opaque training, and insufficient evaluation. The full workflow is shown in Figure~\ref{fig:framework}.

\subsection{Data governance}
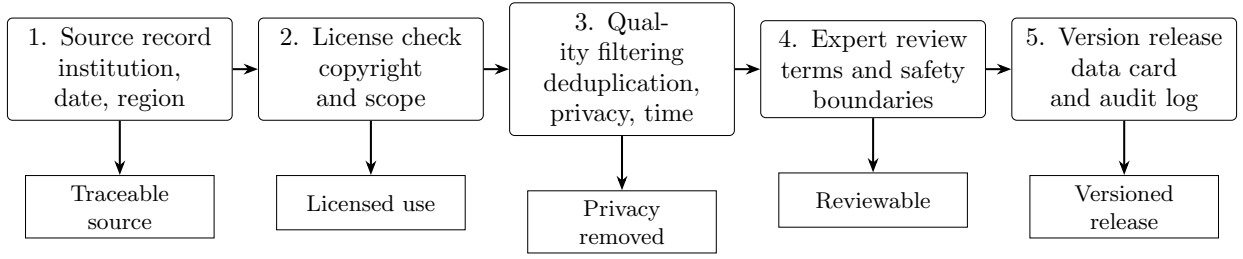
\begin{figure}[htbp]
\centering
\resizebox{0.98\textwidth}{!}{%
\begin{tikzpicture}[
    gstep/.style={draw, rounded corners=2pt, align=center, text width=2.85cm, minimum height=1.0cm, font=\small},
    note/.style={draw, align=center, text width=2.35cm, minimum height=0.75cm, font=\footnotesize},
    arrow/.style={-{Stealth[length=2mm]}, thick}
]
\node[gstep] (s1) {1. Source record\\institution, date, region};
\node[gstep, right=0.35cm of s1] (s2) {2. License check\\copyright and scope};
\node[gstep, right=0.35cm of s2] (s3) {3. Quality filtering\\deduplication, privacy, time};
\node[gstep, right=0.35cm of s3] (s4) {4. Expert review\\terms and safety boundaries};
\node[gstep, right=0.35cm of s4] (s5) {5. Version release\\data card and audit log};
\draw[arrow] (s1) -- (s2); \draw[arrow] (s2) -- (s3); \draw[arrow] (s3) -- (s4); \draw[arrow] (s4) -- (s5);
\node[note, below=0.75cm of s1] (n1) {Traceable source};
\node[note, below=0.75cm of s2] (n2) {Licensed use};
\node[note, below=0.75cm of s3] (n3) {Privacy removed};
\node[note, below=0.75cm of s4] (n4) {Reviewable};
\node[note, below=0.75cm of s5] (n5) {Versioned release};
\draw[arrow] (s1) -- (n1); \draw[arrow] (s2) -- (n2); \draw[arrow] (s3) -- (n3); \draw[arrow] (s4) -- (n4); \draw[arrow] (s5) -- (n5);
\node[align=center, font=\footnotesize, below=0.75cm of n3] {Each sample stores source, date, license status, task type, evidence passage, and review record.};
\end{tikzpicture}%
}
\caption{Agricultural data governance and sample construction pipeline.}
\label{fig:governance}
\end{figure}

Agricultural data should be drawn from authoritative, licensed, and traceable sources, such as government documents, extension manuals, crop-cultivation textbooks, pesticide labels and registration rules, agricultural standards, expert-reviewed QA, and properly licensed datasets. Each sample must keep source, publication date, license status, task type, evidence passage, and review record. The governance pipeline is shown in Figure~\ref{fig:governance}.

\begin{table}[htbp]
\centering
\caption{Admission and exclusion rules for agricultural data sources}
\begin{tabularx}{\textwidth}{p{3cm}X p{3.2cm}}
\toprule
Data type & Admission requirement & Exclusion condition \\
\midrule
Policy documents & Clear source, date, and region of applicability & Expired policy or unverifiable publisher \\
Extension materials & Published by agricultural research, extension, or educational institutions & Anonymous, unsourced, or advertisement-like content \\
Pest and disease material & Includes crop, symptoms, occurrence conditions, and control principles & Image labels without diagnostic rationale \\
Pesticide and fertilizer information & Based on registration, labels, standards, or official instructions & Content encouraging overdose or unverifiable dosage \\
Expert QA & Expert identity and review record can be documented, and answers have evidence links & Personal data are not anonymized or content cannot be reviewed \\
\bottomrule
\end{tabularx}
\end{table}

\subsection{Task and instruction construction}
Agricultural instruction data should contain not only a question and an answer, but also task type, crop, region, growth stage, evidence, risk level, and uncertainty notes. A recommended sample structure is:
\begin{quote}
\small
\textbf{instruction}: Explain the typical symptoms and control principles of rice blast based on the evidence.\\
\textbf{context}: Crop=rice; region=South China; growth stage=tillering; evidence=Section X of an extension manual.\\
\textbf{response}: Describe symptoms, possible causes, field-confirmation steps, control principles, and cases requiring local expert consultation.\\
\textbf{safety}: Do not provide off-label pesticide dosages; do not replace an in-person plant-protection diagnosis.
\end{quote}

\subsection{Parameter-efficient fine-tuning plan}
Let the pretrained weights be $W_0$. LoRA represents a weight update as a low-rank decomposition:
\begin{equation}
W = W_0 + \Delta W = W_0 + BA,
\end{equation}
where $A \in \mathbb{R}^{r \times d}$, $B \in \mathbb{R}^{d \times r}$, and $r$ is the rank. During training, $W_0$ is frozen and only the low-rank matrices are optimized. If QLoRA is used, the base model is stored in quantized form while gradients are propagated through LoRA adapters. We recommend Qwen3-8B as the base model for future empirical experiments and require all training configurations to be disclosed in the model card.

\begin{figure}[htbp]
\centering
\resizebox{0.98\textwidth}{!}{%
\begin{tikzpicture}[
    block/.style={draw, rounded corners=2pt, align=center, minimum height=0.88cm, text width=2.25cm, font=\small},
    adapter/.style={draw, rounded corners=2pt, align=center, minimum height=0.78cm, text width=2.05cm, font=\footnotesize, fill=black!6},
    arrow/.style={-{Stealth[length=2mm]}, thick},
    dashedline/.style={dashed, thick}
]
\node[block] (input) {Agricultural instruction\\evidence and risk tags};
\node[block, right=0.58cm of input] (tok) {Context packing\\input and target};
\node[block, right=0.58cm of tok] (base) {Frozen base\\Qwen3-8B};
\node[block, right=0.58cm of base] (loss) {Supervised loss\\adapter update};
\node[block, right=0.58cm of loss] (save) {Agricultural adapter\\versioned release};
\node[adapter, above=0.62cm of base] (attn) {Trainable LoRA\\Attention};
\node[adapter, below=0.62cm of base] (mlp) {Trainable LoRA\\MLP/FFN};
\draw[arrow] (input) -- (tok);
\draw[arrow] (tok) -- (base);
\draw[arrow] (base) -- (loss);
\draw[arrow] (loss) -- (save);
\draw[dashedline] (attn.south) -- (base.north);
\draw[dashedline] (mlp.north) -- (base.south);
\end{tikzpicture}%
}
\caption{LoRA/QLoRA adaptation structure based on Qwen3-8B. Gray modules denote trainable adapters; the figure describes the training structure and does not claim completed training.}
\label{fig:lora}
\end{figure}
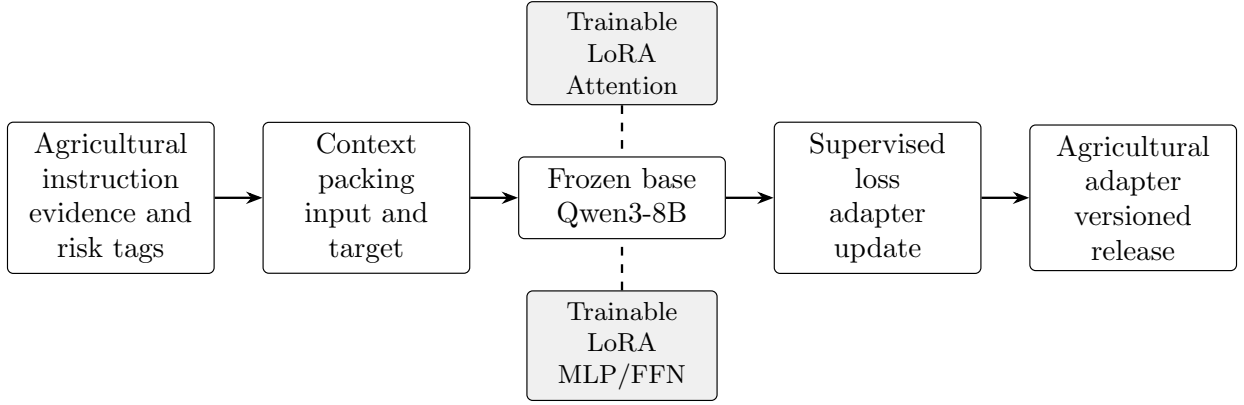

Figure~\ref{fig:lora} shows the recommended LoRA/QLoRA adaptation structure and clarifies which modules should be recorded as trainable or frozen in a future empirical run.

\begin{table}[htbp]
\centering
\caption{Recommended configuration records for reproducible empirical fine-tuning}
\begin{tabularx}{\textwidth}{p{3.3cm}X}
\toprule
Configuration item & Recommendation and required record \\
\midrule
Base model & Qwen3-8B; record exact weight version, download time, license, and checksum information. \\
Adaptation method & LoRA or QLoRA; record rank $r$, alpha, dropout, and target modules. \\
Training data & Use only governed and licensed data; record data version, sample count, and deduplication procedure. \\
Validation set & Stratify by crop, task, region, and risk level to avoid training-set leakage. \\
Evaluation & Automatic metrics are auxiliary; conclusions must include agricultural expert scoring and evidence-consistency checking. \\
Randomness control & Record seed, framework version, hardware, batch size, learning rate, and epoch count. \\
\bottomrule
\end{tabularx}
\end{table}

\subsection{Retrieval-augmented generation}
The agricultural knowledge base should be maintained as an independent module. Each entry should include the original text, source, publication date, applicable region, topic tags, and citable passages. For high-risk agricultural questions, the model should answer based on retrieved evidence. If evidence is insufficient, the model should express uncertainty or recommend consultation with local agricultural experts. RAG is not a guarantee of correctness: retrieval recall, evidence ranking, and citation quality must be evaluated separately.

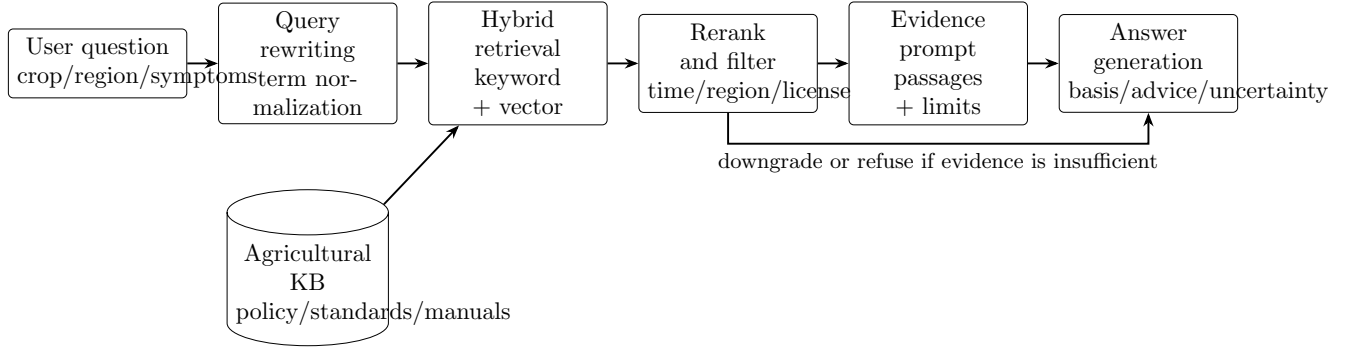
\begin{figure}[htbp]
\centering
\resizebox{0.98\textwidth}{!}{%
\begin{tikzpicture}[
    box/.style={draw, rounded corners=2pt, align=center, minimum height=0.92cm, text width=2.35cm, font=\small},
    store/.style={draw, cylinder, shape border rotate=90, aspect=0.28, align=center, minimum height=1.0cm, text width=2.1cm, font=\small},
    arrow/.style={-{Stealth[length=2mm]}, thick}
]
\node[box] (q) {User question\\crop/region/symptoms};
\node[box, right=0.45cm of q] (rewrite) {Query rewriting\\term normalization};
\node[store, below=0.95cm of rewrite] (kb) {Agricultural KB\\policy/standards/manuals};
\node[box, right=0.45cm of rewrite] (ret) {Hybrid retrieval\\keyword + vector};
\node[box, right=0.45cm of ret] (rank) {Rerank and filter\\time/region/license};
\node[box, right=0.45cm of rank] (prompt) {Evidence prompt\\passages + limits};
\node[box, right=0.45cm of prompt] (ans) {Answer generation\\basis/advice/uncertainty};
\draw[arrow] (q) -- (rewrite);
\draw[arrow] (rewrite) -- (ret);
\draw[arrow] (kb) -- (ret);
\draw[arrow] (ret) -- (rank);
\draw[arrow] (rank) -- (prompt);
\draw[arrow] (prompt) -- (ans);
\draw[arrow] (rank.south) -- ++(0,-0.45) -| node[pos=0.25,below,font=\footnotesize]{downgrade or refuse if evidence is insufficient} (ans.south);
\end{tikzpicture}%
}
\caption{Agricultural retrieval-augmented generation pipeline. The figure emphasizes evidence selection and uncertainty handling rather than performance.}
\label{fig:rag_pipeline}
\end{figure}

Figure~\ref{fig:rag_pipeline} further specifies the evidence flow of agricultural RAG. Source filtering and evidence-sufficiency checks should be completed before generation.

\section{Evaluation Protocol}
Agricultural LLM evaluation should not rely only on generic QA metrics. This paper divides evaluation into five tasks: agricultural knowledge QA, pest and disease consultation, cultivation management, policy explanation, and high-risk refusal. Each task requires a different set of dimensions, as shown in Table~\ref{tab:eval_matrix}.

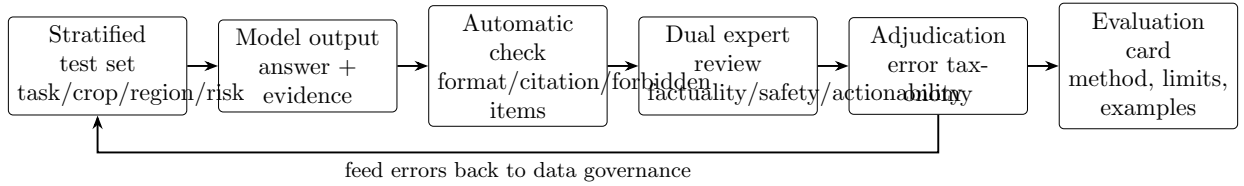
\begin{figure}[htbp]
\centering
\resizebox{0.98\textwidth}{!}{%
\begin{tikzpicture}[
    evalbox/.style={draw, rounded corners=2pt, align=center, minimum height=0.95cm, text width=2.35cm, font=\small},
    arrow/.style={-{Stealth[length=2mm]}, thick}
]
\node[evalbox] (set) {Stratified test set\\task/crop/region/risk};
\node[evalbox, right=0.45cm of set] (out) {Model output\\answer + evidence};
\node[evalbox, right=0.45cm of out] (auto) {Automatic check\\format/citation/forbidden items};
\node[evalbox, right=0.45cm of auto] (expert) {Dual expert review\\factuality/safety/actionability};
\node[evalbox, right=0.45cm of expert] (judge) {Adjudication\\error taxonomy};
\node[evalbox, right=0.45cm of judge] (card) {Evaluation card\\method, limits, examples};
\draw[arrow] (set) -- (out);
\draw[arrow] (out) -- (auto);
\draw[arrow] (auto) -- (expert);
\draw[arrow] (expert) -- (judge);
\draw[arrow] (judge) -- (card);
\draw[arrow] (judge.south) -- ++(0,-0.55) -| node[pos=0.25,below,font=\footnotesize]{feed errors back to data governance} (set.south);
\end{tikzpicture}%
}
\caption{Agricultural model-evaluation loop for human review and error analysis after real experiments.}
\label{fig:evaluation_loop}
\end{figure}

After a real experiment is completed, evaluation should not stop at automatic metrics. Figure~\ref{fig:evaluation_loop} gives a closed loop for expert review, disagreement adjudication, and data improvement.

\begin{table}[htbp]
\centering
\caption{Coverage of tasks and required evaluation dimensions. The symbol $\checkmark$ indicates required evaluation, not a model score.}
\label{tab:eval_matrix}
\small
\begin{tabularx}{\textwidth}{lcccccc}
\toprule
Task & Accuracy & Evidence & Completeness & Actionability & Safety & Uncertainty \\
\midrule
Agricultural QA & $\checkmark$ & $\checkmark$ & $\checkmark$ &  & $\checkmark$ & $\checkmark$ \\
Pest/disease consultation & $\checkmark$ & $\checkmark$ & $\checkmark$ & $\checkmark$ & $\checkmark$ & $\checkmark$ \\
Cultivation management & $\checkmark$ & $\checkmark$ & $\checkmark$ & $\checkmark$ & $\checkmark$ & $\checkmark$ \\
Policy explanation & $\checkmark$ & $\checkmark$ & $\checkmark$ & $\checkmark$ & $\checkmark$ & $\checkmark$ \\
Safety refusal &  & $\checkmark$ &  &  & $\checkmark$ & $\checkmark$ \\
\bottomrule
\end{tabularx}
\end{table}

\begin{table}[htbp]
\centering
\caption{Expert-review rubric}
\begin{tabularx}{\textwidth}{p{2.8cm}X p{2.2cm}}
\toprule
Dimension & Scoring basis & Score \\
\midrule
Accuracy & Whether the answer is consistent with agricultural science, policy, or standards & 1--5 \\
Evidence consistency & Whether the answer is directly supported by retrieved evidence & 1--5 \\
Completeness & Whether the answer covers causes, rationale, actions, and limitations & 1--5 \\
Actionability & Whether the advice is clear, executable, and understandable to farmers & 1--5 \\
Safety & Whether the answer avoids unsafe dosages, illegal pesticides, misleading diagnosis, and absolute promises & 1--5 \\
Uncertainty expression & Whether the answer states boundaries when evidence is insufficient or in-person confirmation is needed & 1--5 \\
\bottomrule
\end{tabularx}
\end{table}

\begin{table}[htbp]
\centering
\caption{Recommended composition of the test set}
\begin{tabularx}{\textwidth}{p{3cm}p{3.5cm}X}
\toprule
Task & Source & Construction requirement \\
\midrule
Agricultural knowledge QA & Textbooks, standards, extension manuals & Answers must have evidence passages and should not rely only on common sense. \\
Pest and disease consultation & Plant-protection material and expert QA & Include symptoms, crop, growth stage, region, and information requiring in-person confirmation. \\
Cultivation management & Cultivation manuals and local extension material & Distinguish general principles from local conditions; avoid absolute yield promises. \\
Policy explanation & Official policy documents & Mark applicable period, region, and responsible authority. \\
Safety refusal & Questions involving illegal dosage, unsafe operations, or unsupported diagnosis & Check whether the model refuses safely and gives safer alternatives. \\
\bottomrule
\end{tabularx}
\end{table}

\section{Safety Control and Responsibility Boundaries}
\begin{figure}[htbp]
\centering
\resizebox{0.98\textwidth}{!}{%
\begin{tikzpicture}[
    box/.style={draw, rounded corners=2pt, align=center, text width=2.55cm, minimum height=1.0cm, font=\small},
    decision/.style={draw, diamond, aspect=2.1, align=center, text width=2.35cm, inner sep=1pt, font=\small},
    arrow/.style={-{Stealth[length=2mm]}, thick}
]
\node[box] (q) {User question\\crop, region, symptoms};
\node[box, right=0.55cm of q] (risk) {Risk detection\\pesticide, diagnosis, policy};
\node[box, right=0.55cm of risk] (retr) {Evidence retrieval\\standards, labels, manuals};
\node[decision, right=0.65cm of retr] (judge) {Sufficient evidence?};
\node[box, above right=0.60cm and 0.7cm of judge] (answer) {Evidence-grounded answer\\with boundaries};
\node[box, below right=0.60cm and 0.7cm of judge] (refuse) {Refuse or escalate\\safer alternative};
\node[box, right=2.55cm of judge] (log) {Audit log\\evidence, risk, output};
\draw[arrow] (q) -- (risk); \draw[arrow] (risk) -- (retr); \draw[arrow] (retr) -- (judge);
\draw[arrow] (judge) -- node[above, font=\footnotesize]{yes} (answer);
\draw[arrow] (judge) -- node[below, font=\footnotesize]{no} (refuse);
\draw[arrow] (answer) -- (log); \draw[arrow] (refuse) -- (log);
\end{tikzpicture}%
}
\caption{Safety-control workflow for high-risk agricultural questions.}
\label{fig:safety}
\end{figure}
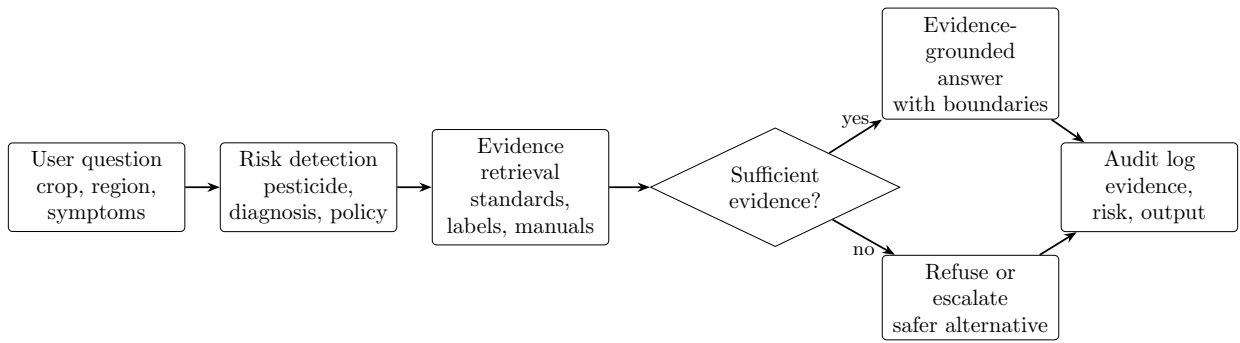

High-risk agricultural questions include pesticide dosage, disease misdiagnosis, food safety, policy eligibility, extreme-weather response, and economic input decisions. A model must not replace the final judgment of local extension experts, plant-protection personnel, or government authorities. High-risk questions should follow risk detection, evidence retrieval, evidence-sufficiency judgment, answer generation, refusal or human escalation, and audit logging, as shown in Figure~\ref{fig:safety}.

\begin{table}[htbp]
\centering
\caption{Handling rules for high-risk agricultural questions}
\begin{tabularx}{\textwidth}{p{3cm}X X}
\toprule
Risk type & The model should do & The model should not do \\
\midrule
Pesticide use & Cite labels, standards, or official material and remind users to follow local rules & Recommend off-label dosage or unclear mixtures \\
Disease diagnosis & Give possibilities, checking steps, and advice for local confirmation & Make an absolute diagnosis based only on text \\
Fertilization and investment & Note the need for soil tests, crop stage, and local conditions & Promise a fixed yield or profit increase \\
Policy consultation & Explain policy text and identify authority and timing conditions & Make a final eligibility decision on behalf of authorities \\
Food safety & Answer conservatively and recommend contacting regulators or professionals & Provide advice for evading regulation or illegal handling \\
\bottomrule
\end{tabularx}
\end{table}

\begin{figure}[htbp]
\centering
\resizebox{0.98\textwidth}{!}{%
\begin{tikzpicture}[
    box/.style={draw, rounded corners=2pt, align=center, minimum height=0.95cm, text width=2.35cm, font=\small},
    sys/.style={draw, rounded corners=2pt, align=center, minimum height=0.95cm, text width=2.35cm, font=\small, fill=black!5},
    arrow/.style={-{Stealth[length=2mm]}, thick}
]
\node[box] (user) {Farmer/extension user\\mobile or web};
\node[sys, right=0.45cm of user] (gateway) {Input gateway\\anonymization and region};
\node[sys, right=0.45cm of gateway] (risk) {Risk classifier\\ordinary/high-risk};
\node[sys, above right=0.65cm and 0.45cm of risk] (rag) {RAG service\\evidence retrieval};
\node[sys, below right=0.65cm and 0.45cm of risk] (human) {Human escalation\\expert or local authority};
\node[sys, right=2.55cm of risk] (model) {Qwen3-8B + adapter\\controlled generation};
\node[sys, right=0.45cm of model] (guard) {Output guardrail\\safety and citation check};
\node[box, right=0.45cm of guard] (reply) {Auditable answer\\advice/evidence/limits};
\draw[arrow] (user) -- (gateway);
\draw[arrow] (gateway) -- (risk);
\draw[arrow] (risk) -- (rag);
\draw[arrow] (risk) -- (human);
\draw[arrow] (rag) -- (model);
\draw[arrow] (human) -- (guard);
\draw[arrow] (model) -- (guard);
\draw[arrow] (guard) -- (reply);
\draw[arrow] (reply.south) -- ++(0,-0.45) -| node[pos=0.35,below,font=\footnotesize]{logging and review} (gateway.south);
\end{tikzpicture}%
}
\caption{Deployment and responsibility boundaries for an agricultural QA system. The diagram emphasizes auditability, privacy, and human escalation.}
\label{fig:deployment}
\end{figure}
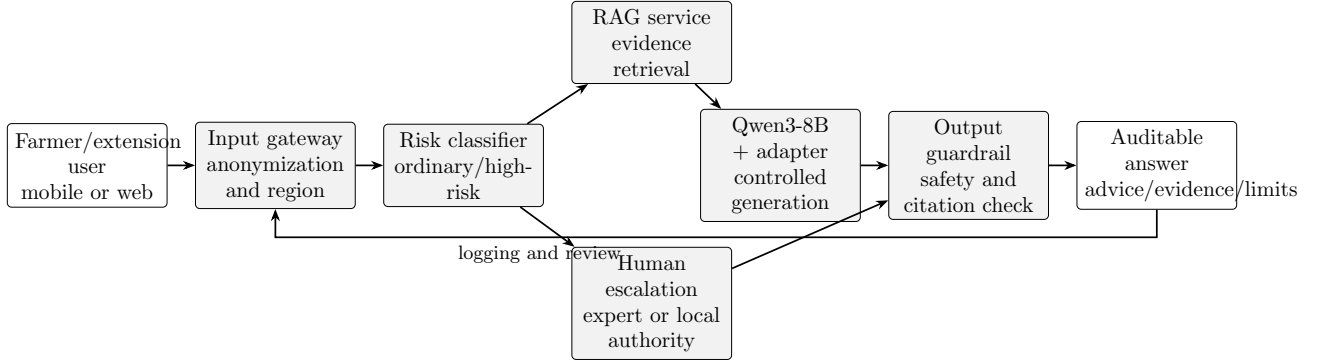

From a system-deployment perspective, an agricultural QA system should include input anonymization, risk classification, retrieval service, output guardrails, and human escalation. Figure~\ref{fig:deployment} shows these responsibility boundaries.

\section{Reproducibility Plan}
To make future empirical studies reproducible, researchers should release or submit the following materials with their model or paper:
\begin{enumerate}[leftmargin=2em]
    \item A data card describing data source, license, time range, region coverage, anonymization, and exclusion rules.
    \item A model card describing base model, adaptation method, training configuration, risk boundary, and known limitations.
    \item An evaluation card describing test-set construction, number of experts, inter-rater agreement, evidence consistency, and error categories.
    \item An audit log recording high-risk questions, refusal examples, insufficient-evidence cases, and human-review outcomes.
\end{enumerate}

\begin{table}[htbp]
\centering
\caption{Truthfulness checklist before submission or release}
\begin{tabularx}{\textwidth}{p{1.2cm}X p{2cm}}
\toprule
No. & Check item & Required status \\
\midrule
1 & Does the paper contain performance numbers unsupported by real experiments? & No \\
2 & Are the base-model name, version, and source verifiable? & Yes \\
3 & Are data source, license, and time range recorded? & Yes \\
4 & Does the paper distinguish protocol design, planned experiments, and real results? & Yes \\
5 & Do high-risk agricultural recommendations have evidence and safety boundaries? & Yes \\
6 & Are limitations and non-replacement of expert judgment explicitly stated? & Yes \\
\bottomrule
\end{tabularx}
\end{table}

\section{Limitations}
The limitation of this paper is explicit: it does not execute real Qwen3-8B agricultural fine-tuning and does not report model-performance results. Therefore, it cannot prove that one fine-tuning configuration is superior to another. Its value lies in providing a truthful methods framework, data-governance pipeline, and evaluation protocol. Future work should add experimental results based on real agricultural data, real training logs, and expert evaluation, including failure cases and safety boundaries.

\section{Conclusion}
This paper proposes AgriTune-R, a reproducible framework for adapting LLMs to agricultural applications. In contrast to writing simulated data as if it were a complete empirical result, this work follows a truthfulness principle: no numerical performance improvement is reported before real training and expert evaluation are completed. We select publicly verifiable Qwen3-8B as the recommended base model and provide data governance, LoRA/QLoRA adaptation, retrieval-augmented generation, expert scoring, and high-risk safety-control procedures. The paper can serve as a protocol foundation for future empirical agricultural LLM studies and for building reliable, safe, and auditable agricultural QA systems.

\end{document}